\documentclass[conference]{IEEEtran}
\IEEEoverridecommandlockouts
\usepackage{cite}
\usepackage{amsmath,amssymb,amsfonts}
\usepackage{algorithmic}
\usepackage{graphicx}
\usepackage{textcomp}
\usepackage{xcolor}
\usepackage{url}
\usepackage{amsmath,amsthm,amsfonts,amssymb,amscd}
\usepackage[ruled,vlined,linesnumbered]{algorithm2e}
\usepackage{commath}

\usepackage{multicol}
\usepackage{colortbl}
\usepackage{multirow, makecell}

\usepackage{makecell}

\usepackage{tabularx}
\newcolumntype{b}{X}
\newcolumntype{s}{>{\hsize=.5\hsize}X}

\usepackage{tabularx}
    \newcolumntype{L}{>{\raggedright\arraybackslash}X}

\usepackage{subfig}

\def\testclr#1#{\@testclr{#1}}
\def\@testclr#1#2{{\fboxsep\z@\fbox{\colorbox#1{#2}{\phantom{XX}}}}}

\def\BibTeX{{\rm B\kern-.05em{\sc i\kern-.025em b}\kern-.08em
    T\kern-.1667em\lower.7ex\hbox{E}\kern-.125emX}}
\begin{document}

\title{Real-time Wireless Transmitter Authorization: Adapting to Dynamic Authorized Sets with Information Retrieval}

\author{\IEEEauthorblockN{Samurdhi Karunaratne, Samer Hanna, and Danijela Cabric 
\thanks{This work was supported in part by the CONIX Research Center, one of six centers in JUMP, a Semiconductor Research Corporation (SRC) program sponsored by DARPA.}
		}
		
\IEEEauthorblockA{\textit{Electrical and Computer Engineering Department,} \\
\textit{University of California, Los Angeles}\\
samurdhi@ucla.edu, samerhanna@ucla.edu, danijela@ee.ucla.edu }
}

\maketitle

\begin{abstract}
As the Internet of Things (IoT) continues to grow, ensuring the security of systems that rely on wireless IoT devices has become critically important. Deep learning-based passive physical layer transmitter authorization systems have been introduced recently for this purpose, as they accommodate the limited computational and power budget of such devices. These systems have been shown to offer excellent outlier detection accuracies when trained and tested on a fixed authorized transmitter set. However in a real-life deployment, a need may arise for transmitters to be added and removed as the authorized set of transmitters changes. In such cases, the system could experience long down-times, as retraining the underlying deep learning model is often a time-consuming process. In this paper, we draw inspiration from information retrieval to address this problem: by utilizing feature vectors as RF fingerprints, we first demonstrate that training could be simplified to indexing those feature vectors into a database using locality sensitive hashing (LSH). Then we show that approximate nearest neighbor search could be performed on the database to perform transmitter authorization that matches the accuracy of deep learning models, while allowing for more than 100x faster retraining. Furthermore, dimensionality reduction techniques are used on the feature vectors to show that the authorization latency of our technique could be reduced to approach that of traditional deep learning-based systems. 
\end{abstract}


\section{Introduction}

With the rapid proliferation of the Internet of Things (IoT), the task of securing IoT networks has become more challenging. Wireless devices in these networks such as sensors are typically constrained by their power and computational capability, rendering traditional cryptography-based authentication systems unsuitable. To address this, passive Physical Layer Authentication (PLA) has been proposed since it does not impose any overhead on the transmitter\cite{wang_wireless_2016}. To identify transmitters, PLA uses channel state information and fingerprints embedded in transmitted signals due to hardware impairments. 

Typically such an authentication system needs to differentiate among transmitters in the authorized set while rejecting unauthorized transmitters (outliers). Since the unauthorized set is practically infinite, this problem has been posed as an open-set classification, as opposed to closed-set classification where all classes are known. Recently, a number of efforts have evaluated open set classification models based on deep learning (DL) in this regard \cite{riyaz_deep_2018} \cite{hanna_open_2021}. They have become the state-of-the-art in PLA, owing to reaching high accuracy while being reasonably robust in the face of channel variations \cite{hanna_open_2021}. 

To the best of our knowledge, these authentication systems have all been evaluated with a static authorized set, meaning that the authorized set of transmitters was assumed to be fixed during training, testing and deployment. However, in most practical situations,  needs change after deployment, resulting in changes to the authorized set: some authorized transmitters might need to be invalidated while others might need to be added. For example, a malfunctioning sensor in an IoT network might need to be replaced with a new sensor. In such cases, it is critical that the authentication system be adapted quickly to the updated authorized set to avoid long down-times. Despite the existence of efficient strategies for retraining DL models, they are still too time-intensive for critical real-time applications like authorization, especially in situations where high availability is key.

In this paper, we propose to use similarity search techniques used in information retrieval applications for open set transmitter authorization. The neural network (NN) of a DL-based authenticator is used to extract feature vectors from a training dataset consisting of authorized, and possibly unauthorized signals. Using the feature vector of each signal sample as its RF fingerprint, we formulate the task of authenticating a query signal as a nearest-neighbor search over the database of RF fingerprints. Since the inference latency associated with an exact nearest-neighbor search is too prohibitive for real-time authentication, locality-sensitive hashing (LSH) is used to partition the database, allowing for a much faster approximate nearest neighbor (ANN) search to be performed. This authorization scheme by design allows for new authorized transmitters to be added by simply indexing signal samples from those transmitters into the database.  Removing authorized transmitters could be accommodated without requiring any changes to the database. Our results show that the proposed LSH scheme is able to achieve retraining times orders of magnitude lower than DL models, with a negligible impact on outlier detection accuracy and inference latency.

Several previous works have used hashing methods to solve the open-set face recognition problem: in \cite{vareto_towards_2017}, the authors paired LSH with fully-connected neural networks, but their approach differs significantly from ours since the purpose for using LSH was for model selection and not for nearest-neighbor search. The closest approach to ours is in \cite{dong_open-set_2020}, where they used LSH to identify most similar faces and thereby solve open-set face identification. However, neither of these approaches considered a dynamic authorized set.

The rest of the paper is organized as follows: we start by formulating the problem in Section \ref{sec:system_model}. Section \ref{sec:adapting_dl_models} discuses how state-of-the-art DL models could be adapted to changes in the authorized set. Section \ref{sec:lsh} presents our LSH-based authorization scheme. An empirical validation of the proposed methods is included in Section \ref{sec:results}. Section VI concludes the paper.

\providecommand{\mA}{\mathcal{A} }
\providecommand{\mK}{\mathcal{K} }
\providecommand{\mO}{\mathcal{O} }
\providecommand{\mX}{\mathcal{X} }
\providecommand{\mY}{\mathcal{Y} }
\providecommand{\mN}{\mathcal{N} }
\providecommand{\mC}{\mathcal{C} }
\providecommand{\mAh}{\bar{\mathcal{A}} }
\providecommand{\mKh}{\bar{\mathcal{K}} }
\providecommand{\mOh}{\bar{\mathcal{O}} }
\providecommand{\mYh}{\mathcal{\hat{Y}} }
\providecommand{\mAn}{\mathcal{A}_{\mathcal{N}} }
\providecommand{\mAr}{\mathcal{A}_{\mathcal{R}} }

\providecommand{\mAc}{\mathcal{|A|} }
\providecommand{\mKc}{\mathcal{|K|} }
\providecommand{\mOc}{\mathcal{|O|} }
\providecommand{\mYhc}{\mathcal{|\hat{Y}|} }
\providecommand{\mAnc}{|\mathcal{A}_{\mathcal{N}}| }

\section{\label{sec:system_model} System Model and Problem Formulation}

\begin{figure}[t]
    \begin{center}
    \vspace*{18px}
    \includegraphics[width=0.48\textwidth]{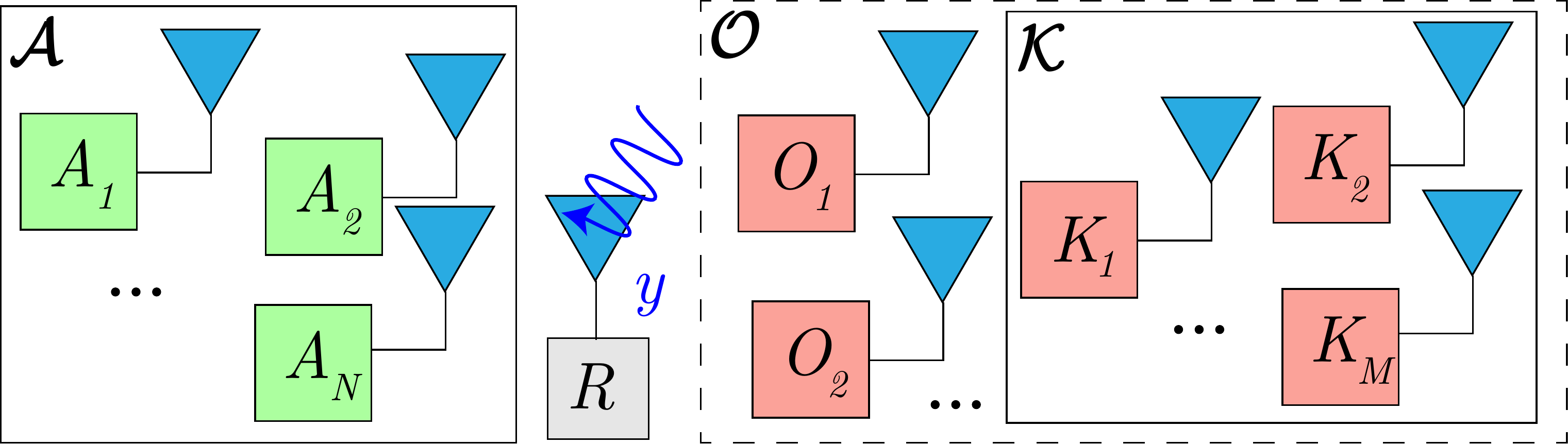}
    \end{center}
    \setlength{\belowcaptionskip}{-5pt} 
    \captionsetup{belowskip=-5pt}
    \caption{System model: $R$ must determine whether the received signal $y$ originated from an authorized transmitter in $\mA$, or from an unauthorized transmitter in $\mO$, some $\mK \subset \mO$ of which may be known to $R$.}%
    \label{fig:system_model}%
\end{figure}

We consider a finite set $\mA=\{A_1,A_2, \cdots,A_\mAc \}$ of transmitters that are authorized to access a system through receiver~$R$. The signal received at $R$ when some transmitter $T$ sends a set of symbols $x$ is $f_T(x)$; $f_T$ models the channel effect, as well as the transmitter fingerprint imprinted on $x$ by $T$ due to the variability of its internal circuitry. The authentication problem can then be formulated as the following binary hypotheses test: based on $y = f_T(x)$, $R$ should determine whether $T$ belongs to the authorized set $\mA$ ($\mathcal{H}_0$) or to the set of outliers $\mO$ ($\mathcal{H}_1$). This is visualized in Fig. \ref{fig:system_model}.

An additional set $\mK=\{K_1,K_2,\cdots,K_\mKc\}$, where $\mK \subset \mO$, of known outliers may be used to improve the outlier detection \cite{hanna_open_2021}. So typically, a dataset of signal samples captured  from transmitters in $\mA$ and a similar dataset captured from transmitters in $\mK$ will be used during training to assist the outlier detector to differentiate between authorized and non-authorized transmitters.

Our task is to adapt to a change in $\mA$ after deploying the authentication system, as quickly as possible. Denote by $\mAh$, $\mKh$ and $\mOh$, respectively, the initial value of $\mA$, $\mK$ and $\mO$. Then, some set $\mAn \in \mOh$ of transmitters could be added to $\mAh$ or some set $\mAr$ could be removed from $\mAh$ and added to $\mOh$ (although both an addition and removal from $\mAh$ could happen, this could be thought of as an addition followed by a removal). 

\section{\label{sec:adapting_dl_models} Adapting deep-learning based classifiers}

In \cite{hanna_open_2021}, we explored several neural network architectures that could be used for the authentication problem such as Disc, DClass and OvA. In this section, we demonstrate how each of these architectures could be adapted to accommodate changes in $\mA$, without entirely retraining the underlying model from scratch.

\begin{table*}[t]
    \centering
    \begin{tabular}{|c|c|c|l}
    \cline{1-3}
     Disc & DClass & OvA & \phantom{dsfsd}\\ \cline{1-3}
     \includegraphics[width=0.15\linewidth]{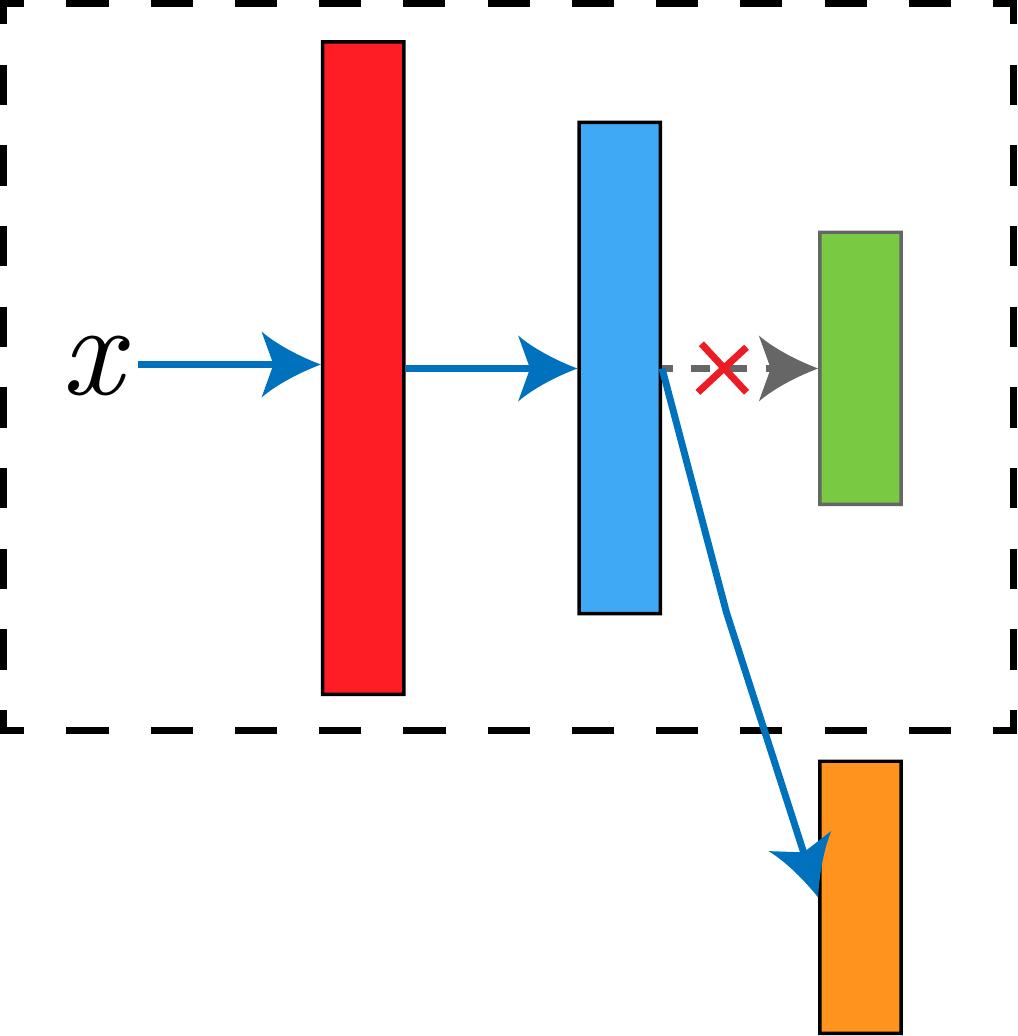}  & \includegraphics[width=0.4\linewidth]{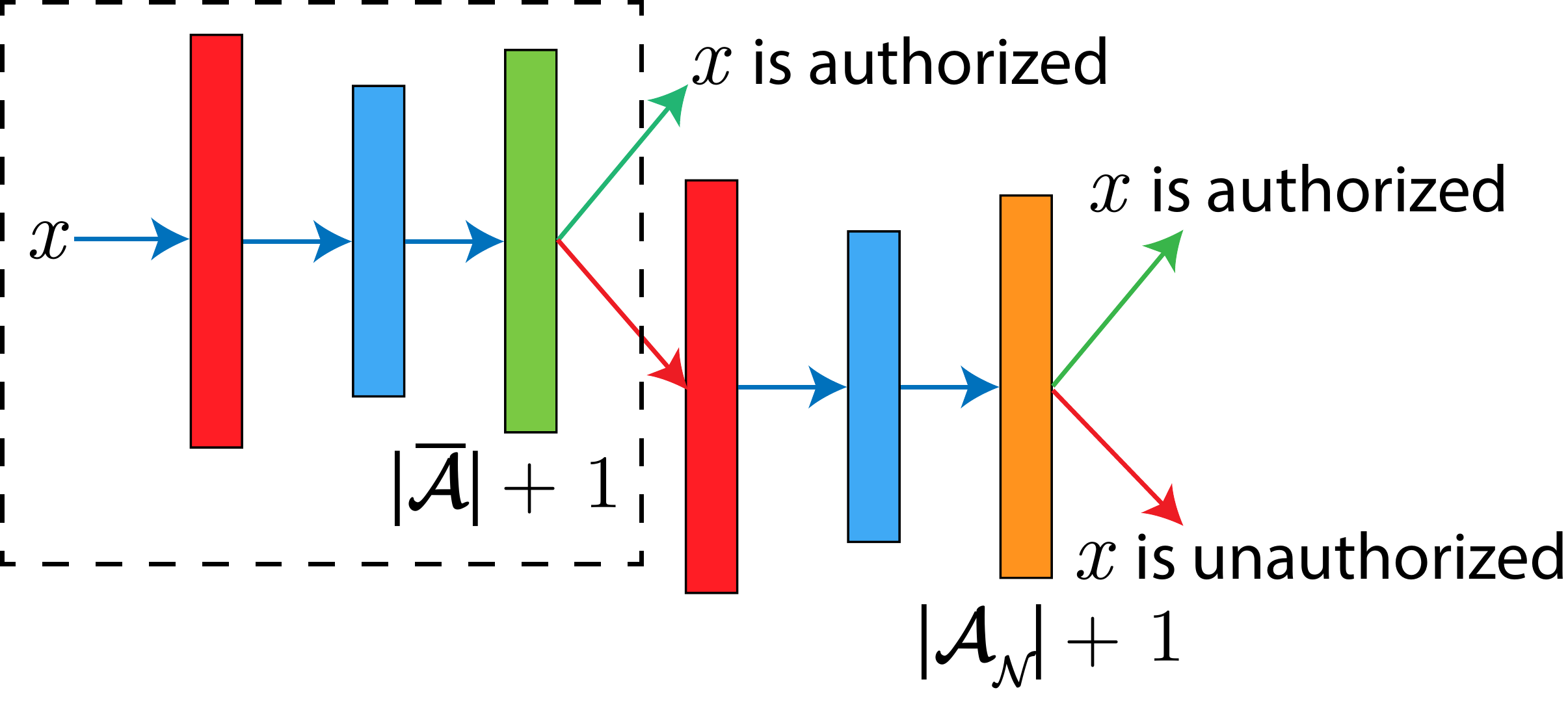}  & \includegraphics[width=0.2\linewidth]{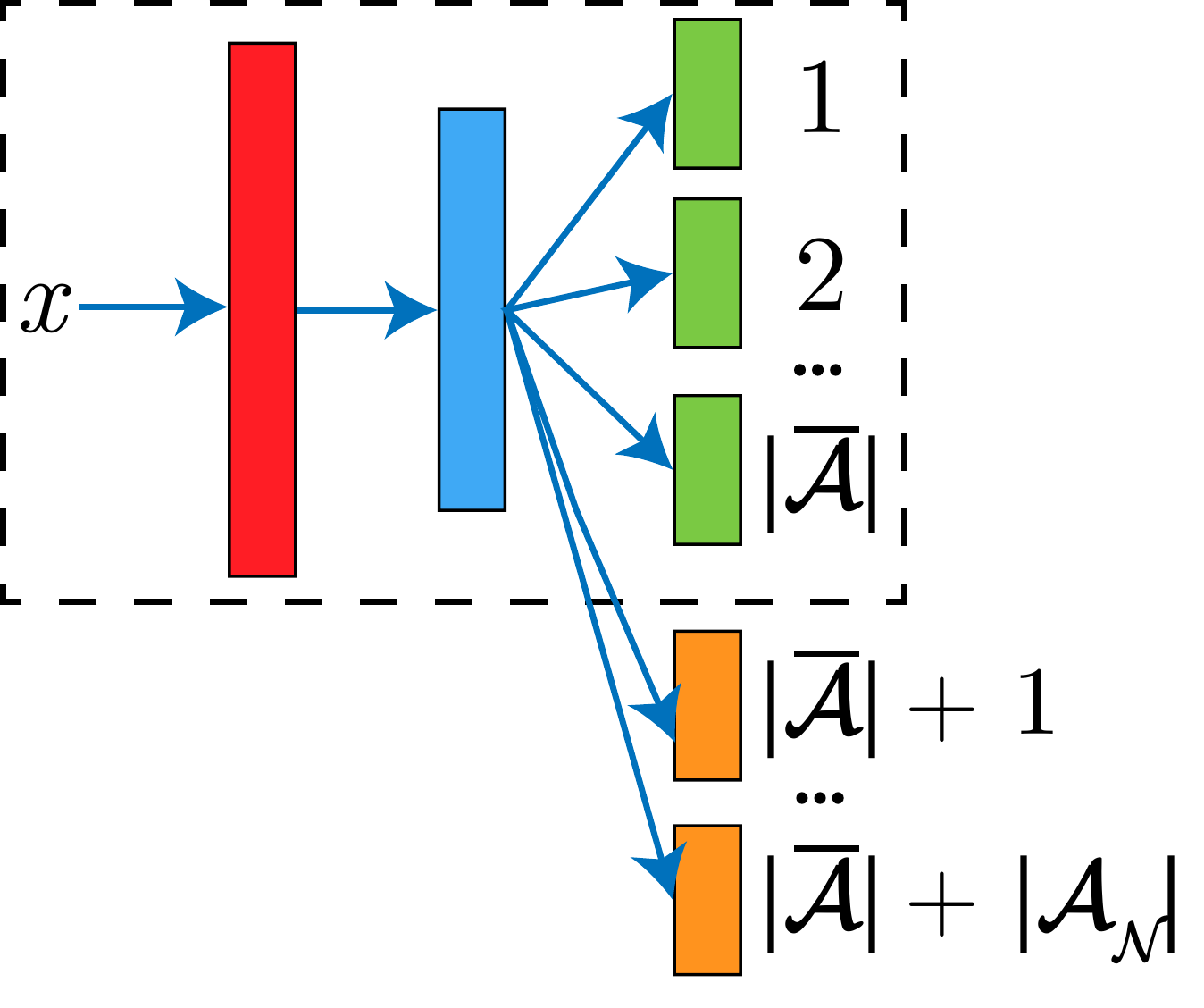} & \includegraphics[width=0.15\linewidth]{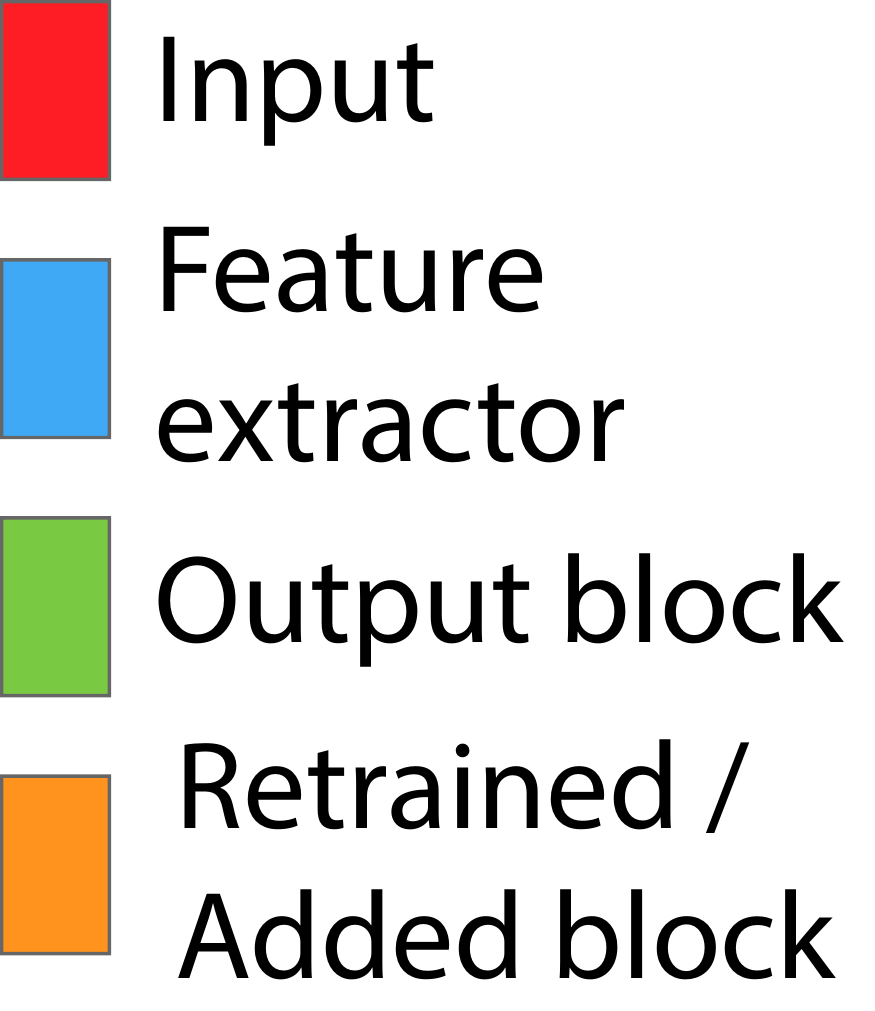}\\ \cline{1-3}
    \end{tabular}
    \caption{Adapting DL models to additions to $\mA$. Dashed-boxes indicate the base model.}
    \label{tab:adding_new_tx}
    \vspace{-4mm}
\end{table*}

The high-level architecture of Disc, DClass and OvA are given in Table \ref{tab:adding_new_tx} (within dashed-boxes), where each could be broken into three building blocks: input, feature extractor and output. The input and feature extractor blocks are similar in all three architectures. In Disc, the output block produces a scalar output through a sigmoid activation indicating its binary authentication decision. OvA has $\mAc$ parallel output blocks, each identical to the output block in Disc, and where the $i$-th block is tasked with independently determining whether the input signal belongs to $A_i$. DClass has one output block with $|\mAh| + 1$ outputs emerging through a softmax activation: the first $|\mAh|$ outputs correspond to authorized transmitters while the last output corresponds to outliers.

Adding transmitters to the authorized set requires a modification of the output block in some form for all three architectures, as summarized in Table \ref{tab:adding_new_tx}. If $\mAn$ is the set of newly added transmitters to $\mAh$, in the case of OvA, this modification could be achieved by adding $\mAnc$ more output blocks in parallel, and retraining the new output blocks while keeping the rest of the NN frozen. Since there is only a single scalar output block in Disc, we could simply retrain that output block. With DClass, a similar approach to Disc is possible, where a new output block with $|\mAh| + \mAnc + 1$ outputs could be trained; however, a more efficient approach would be to utilize the cascaded architecture shown in Table \ref{tab:adding_new_tx}. First we train a secondary network, using $\mAn$ as the authorized set, with the same input and feature extractor blocks as the original network, but with a new output block with $\mAnc + 1$ outputs. A query signal is then judged to be unauthorized only if it is rejected by both NNs. The transmitter-level granularity of OvA and DClass output blocks makes removing transmitters from $\mA$ relatively straightforward: during inference, we simply need to treat the outputs corresponding to the invalidated transmitters as unauthorized. However, Disc does not offer this flexibility, requiring a retraining of the output block as in Table \ref{tab:adding_new_tx}.

Note that except in the case of Disc, it could be potentially very expensive computationally to adapt the NN models to additions to the authorized set, especially for large $\mAnc$, even with the strategies highlighted in Table \ref{tab:adding_new_tx} (we will demonstrate this empirically in Section \ref{sec:results}). This is our motivation to explore alternative authorization schemes that are more adept at efficiently adapting to changes in $\mA$.

\section{\label{sec:lsh} Information retrieval-based transmitter authorization}

\begin{figure*}
    \centering
    \includegraphics[width=1.0\textwidth]{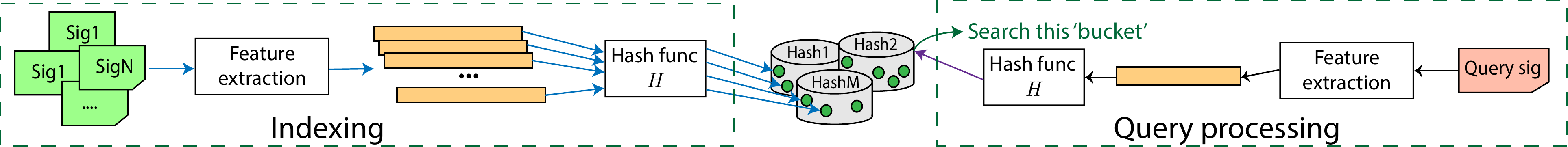}
    \caption{An overview of the indexing and query processing procedure in the LSH authorization scheme}
    \label{fig:lshash_overview}
    \vspace{-4mm}
\end{figure*}

Information retrieval is a broad term that refers to the organization, storage and retrieval of information with respect to a repository of data objects such as signals, documents or images. A typical use case is the task of finding a similar object to a given query object in a repository of objects. More formally, assume we have a repository of $N$ objects, each of dimensionality $d_0$; given a query object $y$, the task is to find an object $y_s$ similar to $y$, based on some similarity metric, as efficiently as possible. In practice, evaluating similarity between objects in the raw data space is ineffective as proximity in data space does not typically correspond to semantic similarity. Therefore, a mapping $y \longrightarrow \hat{y}$ is done from each data object $y \in \mathbb{R}^{d_0}$ to a feature vector $\hat{y} \in \mathbb{R}^{d}$ where the similarity search could be achieved by performing a nearest-neighbor search over a database consisting of those feature vectors.

Assuming we have a training dataset containing sufficient signals from both $\mA$ and $\mK$, a simple algorithm to solve the open set transmitter authorization problem is to find the most similar signal $y_s$ to the query signal $y$: if $y_s \in \mA$ we can infer $y \in \mA$, and $y \in \mO$ if $y_s \in \mK$. A straightforward solution to the similarity problem is to perform an exact nearest neighbor search over the entire database. If the distance between two feature vectors could be computed in $O(d)$ time (e.g. Euclidean distance), this process would take $O(dN)$ time. i.e. linear in $N$.  Assuming that the open set transmitter authorization problem could be solved by performing such a nearest-neighbor search, a per-query linear-time solution is too prohibitive, considering the fact that such an authorization system is expected to serve multiple authorization requests per second. Therefore a sub-linear time search is required.

Approximate nearest-neighbor search algorithms allow us to perform the similarity search in sub-linear time by making the compromise that the returned item need not be the strictly nearest-neighbor, but whose distance to the query object is sufficiently close to that of the strictly nearest-neighbor. A common approach to achieving sub-linearity is to eliminate the need for an exhaustive search by partitioning the database into some $M$ ``buckets" such that $\hat{y}$ and its true nearest neighbor $\hat{y}_s$ are in the same bucket with high probability; then, the exhaustive search for $\hat{y}_s$ need only be done inside that bucket, and not over the entire database. Locality Sensitive Hashing (LSH) \cite{gionis_similarity_1999} could be used to perform the partitioning such that this property holds. 

Cryptographic hash functions (CHFs) attempt to create a large deviation in the hash value when there is a slight deviation in the input; conversely, LSH functions try to create hash values that preserve locality. In particular, LSH functions ensure that inputs that are close in the input space receive the same hash value with high probability. Although there are a number of LSH functions proposed in the literature, in this paper we chose the function $H$ based on random projections, mainly due to its simplicity and ease of implementation. For an input $\hat{y}$, the hash value $H(\hat{y})$ is a binary string calculated as following: $K$ hyperplanes $w_1,w_2,\dots,w_{K}$ are randomly generated where $w_i \in \mathbb{R}^{d}~\forall~i \in \{1, \dots, K\}$; then, the $i$-th bit of $H(\hat{y})$, $[H(\hat{y})]_i$ is set to 1 or 0 depending on whether the point $\hat{y}$ is above or below the hyperplane $w_i$ in $\mathbb{R}^{d}$ space. Here, $K$ is the length of the hash value, called the hash size (note that there are $2^{K}$ possible hash values). With $H$ defined this way, the indexing process is simply to place each signal $x$ in the bucket labeled with hash value $H(\hat{x})$, as visualized in Fig. \ref{fig:lshash_overview}.

\subsection{Using LSH database to perform authorization}

Assume we have indexed a set of training signals $\mathcal{X}$ into an LSH database; $\mathcal{X}$ includes signal samples from $\mA$ and possibly samples from $\mK$. For a query signal $y$, we can use the LSH database to determine whether or not $y \in \mA$ in a two step inference process:
\begin{itemize}
    \item \textbf{Step 1:} Determine $\text{NN}_{\text{LSH}}(\hat{y})$, the approximate nearest-neighbor of $\hat{y}$. If $\text{NN}_{\text{LSH}}(\hat{y})$ does not exist, we infer that $y \notin \mA$. Otherwise, we move to the next step. 
    \item \textbf{Step 2:} Let $\hat{y}_s = \text{NN}_{\text{LSH}}(\hat{y})$. If $y_s \notin \mA$, we infer that $y \notin \mA$, and that $y \in \mA$ otherwise.
\end{itemize}

Note that the existence of $\text{NN}_{\text{LSH}}(\hat{y})$ in Step 1 is not guaranteed since the randomization involved means that similar items are not guaranteed to be grouped correctly. This shortcoming could be overcome by creating $L$ LSH databases instead of one, where the set of $K$ hyperplanes is generated independently in each case. Here, the exact nearest-neighbor search is performed on all buckets mapped to $\hat{y}$ over all $L$ databases, increasing the chance that a nearest-neighbor is found. Furthermore, it should be noted that the two-step process above does not require $\mathcal{X}$ to contain samples from $\mK$; in that case, intuitively $\text{NN}_{\text{LSH}}(\hat{y})$ should not exist as long as $K$ is large enough (there are enough buckets). 

\subsection{Feature extraction}

It has been shown that the activations produced by deeper layers of convolutional neural networks trained for image classification tasks could be used as a high-level image descriptor \cite{babenko_neural_2014}. Inspired by this, we propose to use the activations invoked by the feature extractor block of a trained transmitter authorization NN model as the feature vector for a signal in our LSH authorization scheme. We call this NN our embedding model since it is used to extract a feature vector or embedding. Although this creates a dependence on a standard DL-based classifier, the expectation is that as long as the initial embedding model is expressive enough (trained on a sufficiently large dataset), it does not need to be retrained when the authorized set changes.   

\subsection{Adapting to changes in $\mA$}

With the authorization scheme described above, it is straightforward to adapt to changes in $\mA$. If transmitters in $\mAn$ are added to $\mAh$, then we simply need to index signal samples collected from transmitters in $\mAn$ to the LSH database. If some transmitters $\mAr \subset \mAh$ are removed from $\mAh$, then no modification to the LSH database is necessary: during Step 2 of the inference process, it should simply be noted that if $y_s \in \mAr$, then in fact $y_s \notin \mA$.

\subsection{\label{sec:complexity} Computational complexity and feature vector compression}

It is easy to see that the indexing process has a cost of $NLdK$ (cost of $K$ $d$-dimensional dot products for the $N$ data-points, repeated for all $L$ databases). Since $N = |\mathcal{X}| >> \mAc$, the computational complexity of the two-step inference process is essentially the same as that of the $\text{NN}_{\text{LSH}}(\cdot)$ operation:
\begin{enumerate}
    \item Calculating $H(\cdot)$ has a cost of $dK$ since a $d$-dimensional dot product needs to be calculated $K$ times
    \item If all $N$ data-points are distributed evenly over the $2^K$ buckets, then the exact nearest-neighbor search will constitute calculating the distance metric over $N/2^K$ data-points for a total cost of $d \times N/2^K$.
    \item Since $\text{NN}_{\text{LSH}}(\cdot)$ is evaluated over $L$ databases, the total inference cost is $L \times (dK+dN/2^K)$. 
\end{enumerate}

Note that both the indexing cost and the inference cost has a linear dependence on the dimensionality of the feature vectors $d$. Therefore we could also attempt to add a dimensionality-reduction step during indexing as well as during inference; in this paper, we tested the use of an auto-encoder model for this purpose. Note that similar to the embedding model used for feature-extraction, the encoder does not need to be retrained during changes to $\mA$, as long as the initial auto-encoder was trained on a sufficiently large dataset.

\section{\label{sec:results} Experimental Evaluation}

\begin{table*}[t]
    \centering
    \begin{tabularx}{\linewidth}{
    |>{\hsize=0.6\hsize}X|
    >{\hsize=1.6\hsize}X|
    >{\hsize=0.6\hsize}X|
    >{\hsize=1.2\hsize}X|
  }
         \hline
         Auth. scheme & Description & Trained on & Retrained on\\ \hline
         \textit{DClass} & Initial DClass model & $\mathcal{X}_{train}$ and  $\mathcal{X}_{val}$ & $\mathcal{X}^{\mN}_{train}$ and $\mathcal{X}^{\mN}_{val}$ (adapted as in Table \ref{tab:adding_new_tx})\\ \hline
         \textit{DClass sep} & Initial DClass model retrained from scratch & $\mathcal{X}_{train}$ and  $\mathcal{X}_{val}$ & on $\mathcal{X}^{\mC}_{train}$ and $\mathcal{X}^{\mC}_{val}$ \\ \hline
         \textit{LSH} & Standard LSH scheme & $\mathcal{X}$ & $\mathcal{X}^{\mN}$ \\ \hline
         \textit{LSH small} & LSH scheme with a smaller database & 300 samples from $\mathcal{X}$ & 300 samples from $\mathcal{X}^{\mN}$ \\ \hline
         \textit{LSH dim-red} & LSH scheme with dimensionality-reduced feature vectors & $\mathcal{X}$ & $\mathcal{X}^{\mN}$ \\ \hline
         \textit{LSH dim-red small} & Similar to \textit{LSH dim-red} but with a smaller database & 300 samples from $\mathcal{X}$ & 300 samples from $\mathcal{X}^{\mN}$ \\ \hline
    \end{tabularx}
    \caption{Different authorization schemes considered in the experimental evaluation}
    \label{tab:auth_schemes}
\end{table*}

We start by introducing the dataset and evaluation procedure, and discuss results obtained for different experiments. 

 \subsection{Dataset \label{sec:dataset}}
A dataset consisting of 71 transmitters was captured on the Orbit testbed \cite{orbit_2005}. The receiver was a software defined radio (USRP N210) and each transmitter was an off-the-shelf Atheros WiFi module allowed to transmit over Channel 11 (with a center frequency of 2462 MHz and bandwidth of 20 MHz). Energy detection was used to extract packets after an IQ capture at a rate of 25 Msps for 1 second. Without any synchronization or further preprocessing, we used the first 256 IQ samples of each packet, containing the preamble, as the signal sample.

\subsection{Evaluation Procedure}

\begin{figure*}[t]
	\centering
	\vspace{-9mm}
	\subfloat[Retraining time \label{fig:vary_num_add_auth_retraining_time}]{\includegraphics[width=0.33\linewidth]{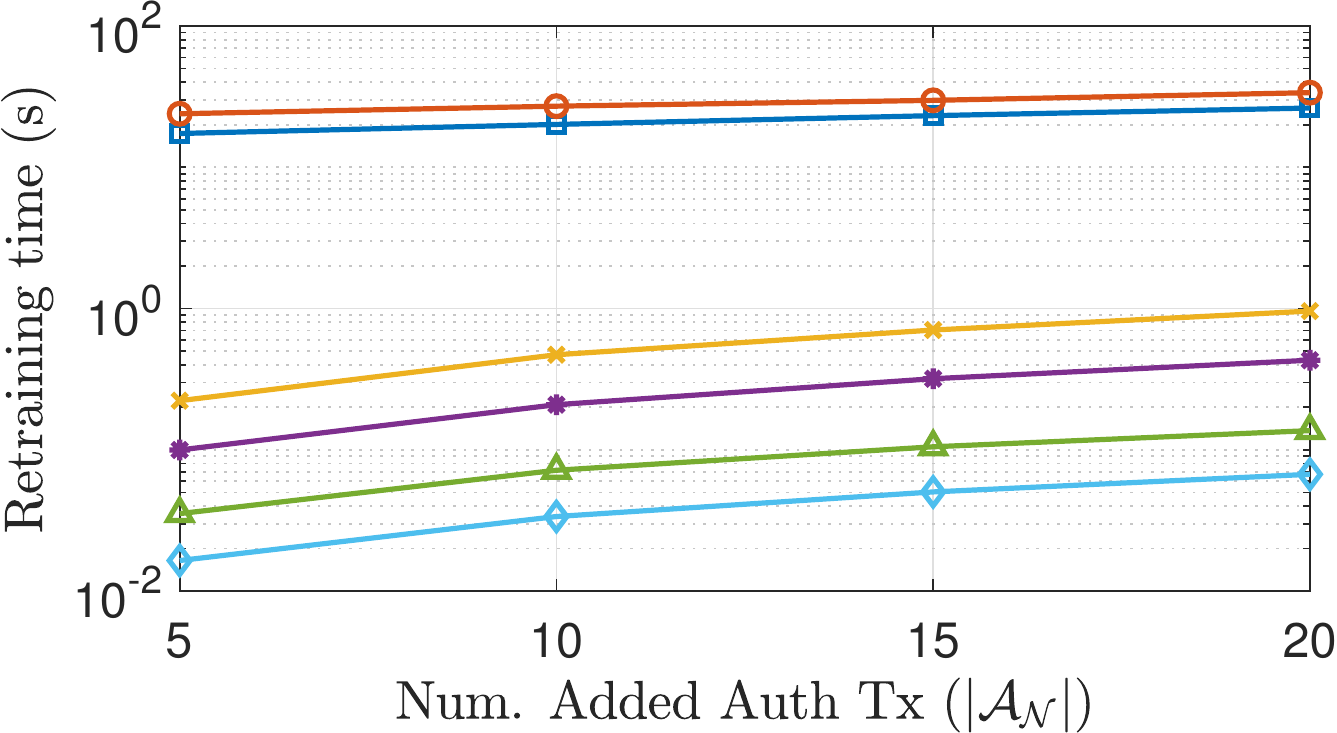}}
	\subfloat[Accuracy \label{fig:vary_num_add_auth_accuracy}]{\includegraphics[width=0.33\linewidth]{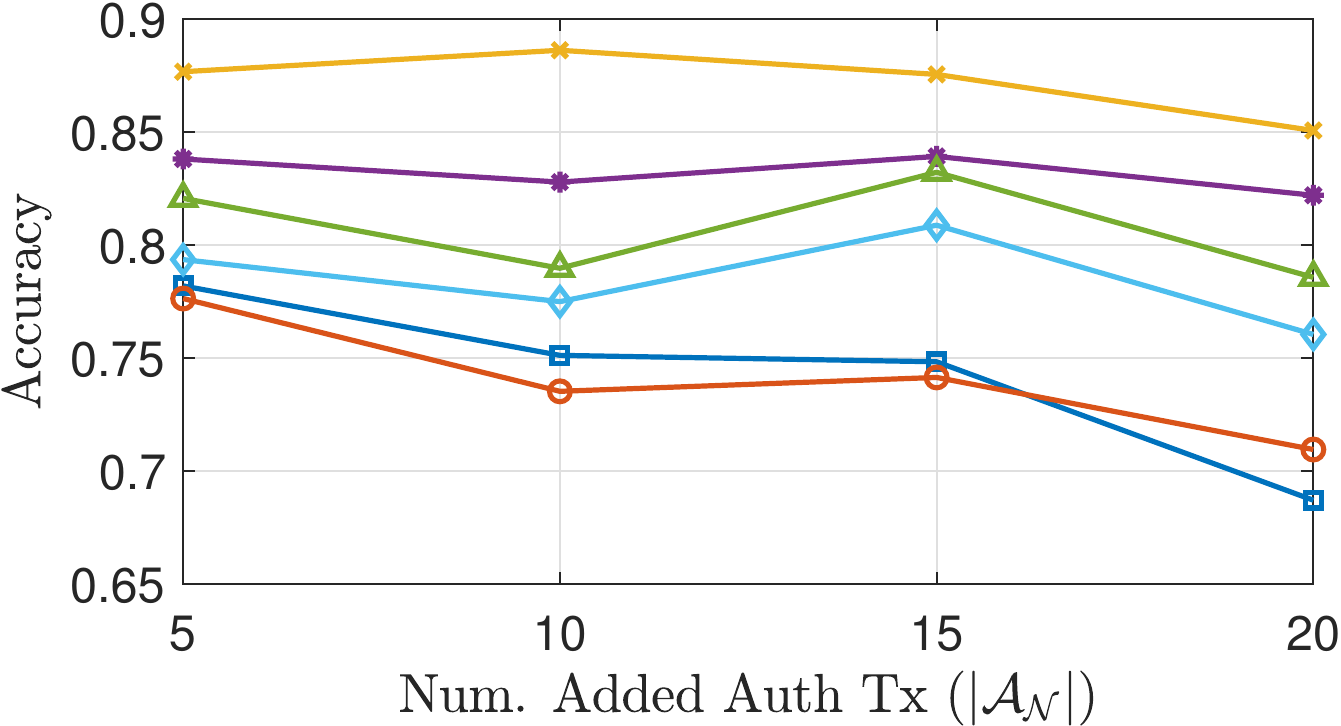}}
	\subfloat[Inference Latency \label{fig:vary_num_add_auth_latency}]{\includegraphics[width=0.33\linewidth]{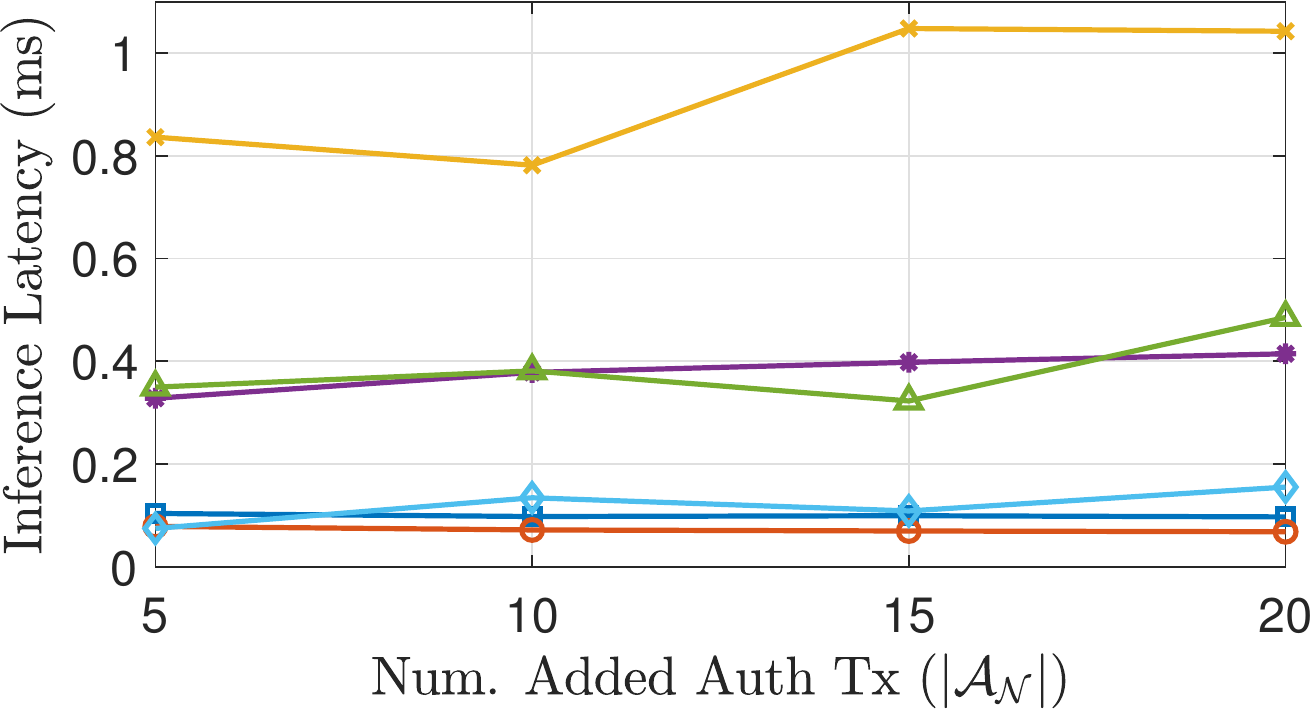}}

	\vspace{-3mm}
	\subfloat{\includegraphics[width=0.8\linewidth]{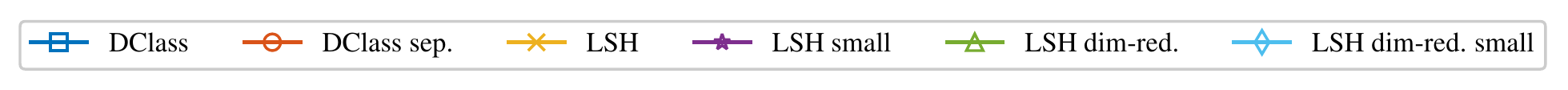}} 
	\caption{Performance of different authorization schemes against $\mAnc$}
	\label{fig:vary_num_add_auth}
	\vspace{-5mm}
\end{figure*}

\begin{figure*}[t]
	\centering
	\subfloat[Accuracy \label{fig:heatmap_accuracy}]{\includegraphics[width=0.24\linewidth]{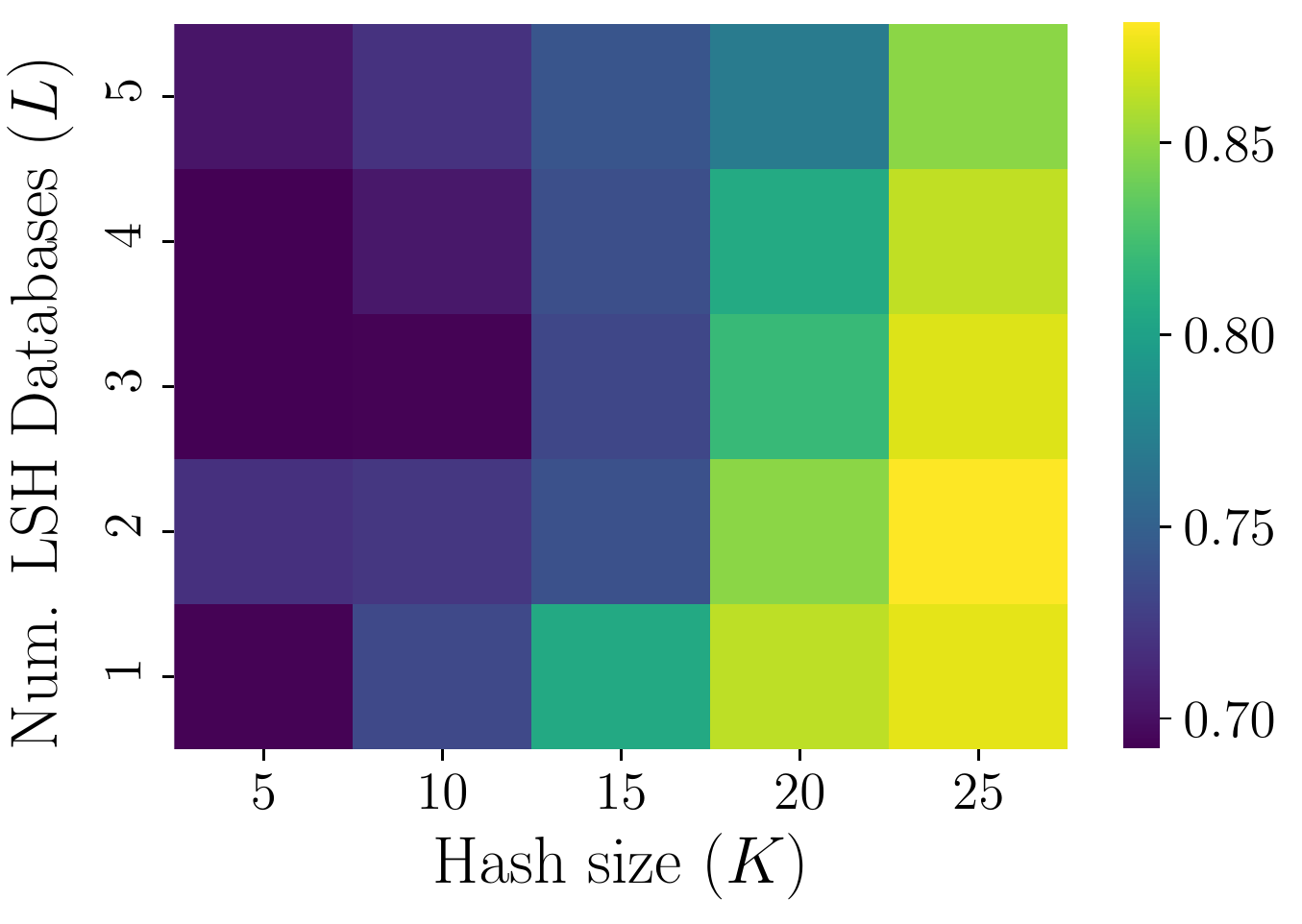}}\hspace{1mm}
	\subfloat[Precision \label{fig:heatmap_precision}]{\includegraphics[width=0.24\linewidth]{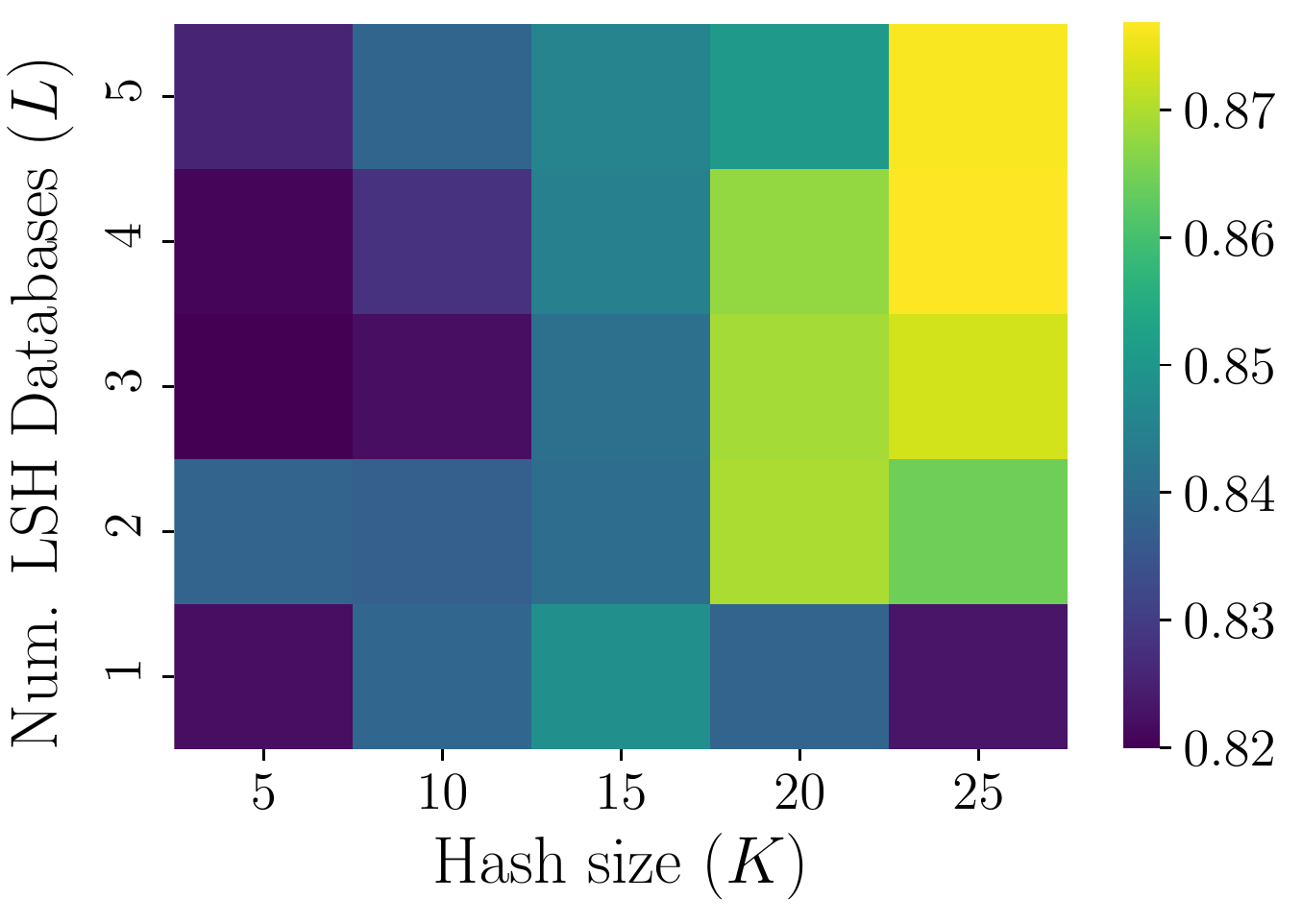}}\hspace{1mm}
	\subfloat[Recall \label{fig:heatmap_recall}]{\includegraphics[width=0.24\linewidth]{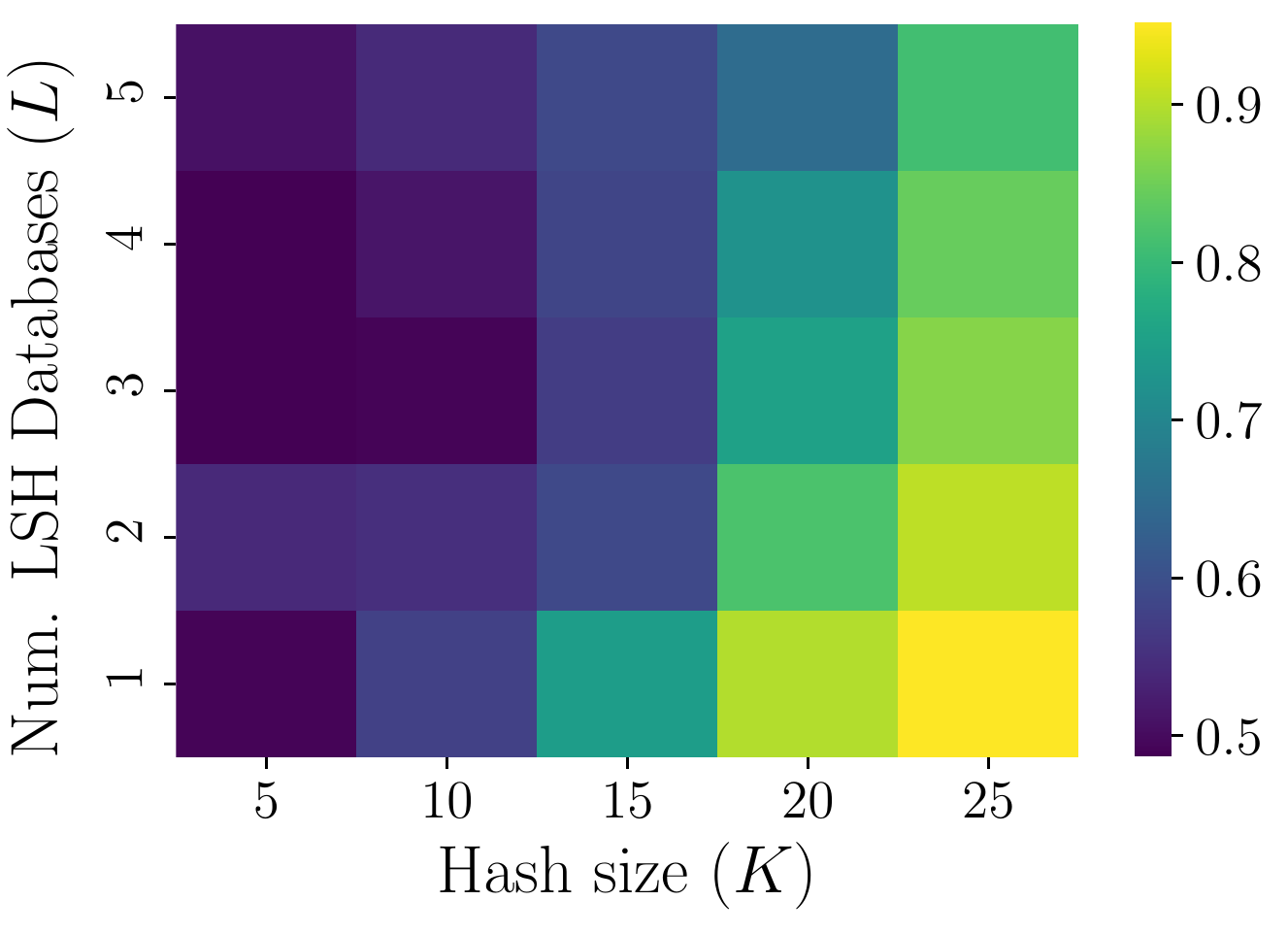}} \hspace{1mm}
	\subfloat[Inference Latency (ms) \label{fig:heatmap_latency}]{\includegraphics[width=0.24\linewidth]{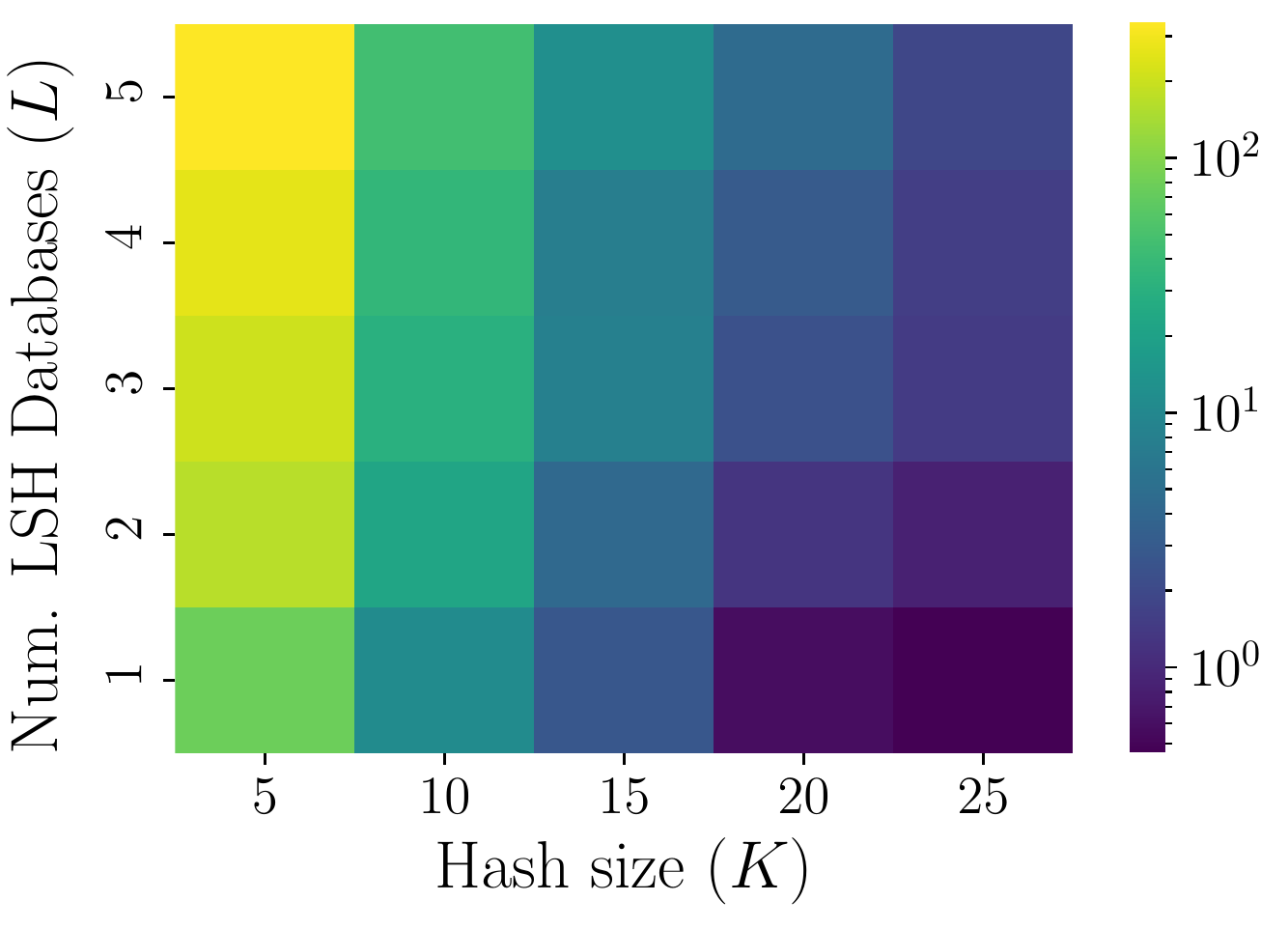}} 
	\caption{Performance of \textit{LSH} with the variation of $L$ and $K$}
	\label{fig:heatmaps}
	\vspace{-5mm}
\end{figure*}

As explained in Section III, removing transmitters from the authorized set is a relatively inexpensive procedure for all the NN architectures in Table \ref{tab:adding_new_tx}; therefore, we will only focus on the case of adding transmitters to the authorized set. Also, we will only use DClass for comparisons with the LSH scheme since it has better outlier detection accuracy than Disc while being less computationally intensive to train than OvA \cite{hanna_open_2021}, offering a more fair comparison.

$\mA$, $\mK$ and $\mO$ will be chosen randomly, subject to the constraints specified for each evaluation---however, when comparing different authorization schemes, the same $\mA$, $\mK$ and $\mO$ will be kept. For chosen $\mA$, $\mK$ and $\mO$, the dataset split will be as follows: for the training dataset $\mathcal{X}_{train}$ and the validation dataset $\mathcal{X}_{val}$, we use 70\% of the samples belonging to $\mA$, and all the samples belonging $\mK$. The shuffled combination of this data is split into 80\% for $\mathcal{X}_{train}$ and 20\% for $\mathcal{X}_{val}$. The test $\mathcal{X}_{test}$ set contains all samples from $\mO$ and the remaining 30\% of $\mA$. We will define this method of splitting the dataset for some $\mA$, $\mK$ and $\mO$ as $\textbf{split}(\mA, \mK, \mO) = \{\mathcal{X}_{train}, \mathcal{X}_{val}, \mathcal{X}_{test}, \mathcal{X}\}$ where $\mathcal{X} = \mathcal{X}_{train} \cup \mathcal{X}_{val}$.

For each $\mA$, $\mK$ and $\mO$, we start with a DClass model trained on $\mathcal{X}_{train}$ and $\mathcal{X}_{val}$, and an LSH authorization scheme where $\mathcal{X}$ is used to create the initial LSH database. The composition of the DClass feature extractor block was the same as that used in \cite{hanna_open_2021}. A frozen copy of the initial DClass model will be used as the embedding model for any LSH authorization schemes. An auto-encoder is also trained on $\mathcal{X}$; the resulting encoder is isolated and frozen to be used as the encoder for any dimensionality reduction. Then for a given value of $\mAnc$, a set of $\mAnc$ transmitters will be randomly chosen from $\mOh$ as $\mAn$ and the dataset will be split again to form $\textbf{split}(\mAn, \mKh, (\mOh-\mAn) \cup (\mAh-\mAn)) = \{\mathcal{X}^{\mN}_{train}, \mathcal{X}^{\mN}_{val}, \mathcal{X}^{\mN}_{test}, \mathcal{X}^{\mN}\}$ and $\textbf{split}(\mAh \cup \mAn, \mKh, \mOh-\mAn) = \{\mathcal{X}^{\mC}_{train}, \mathcal{X}^{\mC}_{val}, \mathcal{X}^{\mC}_{test}, \mathcal{X}^{\mC}\}$.

Table \ref{tab:auth_schemes} details the set of authorization schemes we use in our experiments, including on which datasets they are trained and retrained on. The \textit{small} datasets are considered because the inference cost is positively correlated with $N$, and therefore should help reduce the inference latency. Note that Euclidean distance was used as the distance metric for LSH schemes. 

Different authorization schemes will be evaluated on $\mathcal{X}^{\mC}_{test}$ with respect to: \begin{itemize}
    \item Accuracy: Outlier detection accuracy  on $\mathcal{X}^{\mC}_{test}$
    \item Inference latency: The time to output the authorization decision per query signal, averaged across $\mathcal{X}^{\mC}_{test}$
    \item Retraining time: The total time required to adapt the deployed authorization system to the change in $\mA$. 
\end{itemize}

It should be stressed that as long as the LSH scheme does not significantly compromise the accuracy and inference latency compared to DClass, retraining time is the critical metric of interest. Training time, which is the total time required to train each authorization system, is not analyzed as it is predictably higher for LSH schemes due to the indexing overhead; this is however, a good compromise to make as the training-phase occurs before the deployment of the authorization system.

\subsection{Adding $\mAnc$ transmitters to $\mAc = 10$}

In this experiment, we fix $L=20, K=1$ and start with $\mAc = 10$, $\mKc = 15$, $\mOc=30$ and then add $\mAnc=\{5, 10, 15, 20\}$ transmitters to $\mA$ from $\mO$. The variation of retraining time, outlier detection accuracy and inference latency versus $\mAc$ are given in Fig. \ref{fig:vary_num_add_auth}. Fig. \ref{fig:vary_num_add_auth_retraining_time} provides strong evidence that LSH authorization schemes are able to adapt to the change in the authorized set much faster than the DL models; in particular, we are able to see a roughly 100x improvement in retraining time (note that the time-axis is in logarithmic scale). Furthermore, from Fig. \ref{fig:vary_num_add_auth_accuracy} we can immediately see that LSH schemes are able to match or even outperform the DClass models in terms of outlier detection accuracy. Also note that \textit{DClass} matches the performance of \textit{DClass sep}, justifying the freeze-and-train method proposed in Table \ref{tab:adding_new_tx}. Fig. \ref{fig:vary_num_add_auth_latency} paints a contrasting picture: DClass models are able to perform authorization decisions much faster than the standard LSH scheme. This justifies the purpose of opting to build smaller LSH databases with dimensionality-reduced features. Note in particular that \textit{LSH dim-red small} is able to match the latency performance of DClass while still slightly outperforming it on accuracy performance. Therefore, it is clear that LSH authorization schemes are a viable alternative to DL models, especially when $\mA$ is expected to evolve over the lifetime of the authorization system.

\subsection{Effect of $K$ and $L$}

Understanding the performance impact of the two hyper-parameters $L$ and $K$ (number of LSH databases and hash size) can help design LSH authorization systems to fit individual needs and flexibilities. To evaluate this, we fixed $\mAc = 10$, $\mKc = 15$, $\mOc=30$, $\mAnc=5$ and varied $L \in \{1, 2, 3, 4, 5\}$,  $K \in \{5, 10, 15, 20, 25\}$ to obtain the results in Fig. \ref{fig:heatmaps}. 

Recall from Section \ref{sec:complexity} that the indexing cost is directly proportional to both $L$ and $K$; therefore as expected, we observed that the retraining time grew with both $L$ and $K$ (not displayed in Fig. \ref{fig:heatmaps} for the sake of brevity). More interestingly, from Fig. \ref{fig:heatmap_precision} and Fig. \ref{fig:heatmap_recall} we see that for large $K$, as $L$ is increased, the precision increases but the recall decreases. Increasing $L$ amortizes the effect of bad hyperplane selections, ensuring that true nearest-neighbors ``collide" (fall to the same bucket) on at least one of the $L$ databases. This results in a decrease of false-positives (authorized signals being flagged as unauthorized) and hence an increase in precision, as it prevents Step 1 of the two-step inference process from failing erroneously. However, increasing $L$ also has the side-effect of increasing the probability that an unauthorized signal collides with authorized signals (imagine a case when $\mathcal{X}$ exclusively has samples from $\mA$), thereby increasing false negatives and hence decreasing the recall. Decreasing $K$ increases the likelihood of false collisions resulting in increased false-negatives and hence lower recall as seen in Fig. \ref{fig:heatmap_recall}. However, if $K$ is too high at low $L$, it could result in similar points not colliding, resulting in false positives and hence low precision; this can actually be seen in Fig. \ref{fig:heatmap_precision}, where for $L=1$, the precision increases at first but then decreases. Due to false negatives varying in a larger range (higher range of recall in Fig. \ref{fig:heatmap_recall}) than false positives (lower range of precision in Fig. \ref{fig:heatmap_precision}), it is unsurprising that in Fig. \ref{fig:heatmap_accuracy} the accuracy follows the same trend as the recall.

Arguably the most surprising result in Fig. \ref{fig:heatmaps} is that higher accuracy in Fig. \ref{fig:heatmap_accuracy} does not come at the cost of higher latency in Fig. \ref{fig:heatmap_latency}; in fact, it seems that higher accuracy is attainable with lower latency. Although this might seem counter-intuitive, it is explainable from the inference cost formula we derived in Section \ref{sec:complexity}: $c = L \times (dK+dN/2^K)$. As it dictates, we can clearly see the linear variation of $c$ with $L$ in Fig. \ref{fig:heatmap_latency}. However, the variation of $c$ with $K$ very much depends on the particular value of $N$; in fact, assuming the formula for $c$ holds, it can be theoretically shown that $K = K_m \approx \log_2(N)$ minimizes $c$. In our case, $N \approx 10^4$ so $K_m \approx 13$ should have been ideal, which is seemingly contradicted in Fig. \ref{fig:heatmap_latency} due to the latency continuing to drop as $K$ is increased upto 25. This discrepancy is most likely due to the assumption made in deriving $c$ that data-points in the LSH database are evenly divided across the $2^K$ buckets, which may not be true due to the nature of the data involved. In fact, as $K$ is increased beyond 25, around $K=100$ the latency starts to increase as the cost of calculating $H(\cdot)$ becomes too prohibitive. 

The takeaway from this experiment is that the performance impact of $L$ and $K$ is hard to predict due to dependence on factors like $N$, composition of $\mathcal{X}$ and the nature of the data involved. Therefore, it would be advisable to use a validation split of the dataset to calibrate them to the specific use case.

\section{\label{sec:conclusion} Conclusion}

In this paper, we considered the problem of adapting to a dynamic authorized set in RF transmitter authorization. First, we demonstrated how state-of-the-art DL models could be adapted to changes in the authorized set. Then we described how locality sensitive hashing could be used to facilitate approximate nearest-neighbor search in the realm of information retrieval to solve the transmitter authorization problem by building an LSH database. With this approach, incorporating changes to the authorized set in terms of additions and removals was shown to be manageable with simple changes to the underlying LSH scheme. From empirical results we showed that LSH schemes offers dramatically reduced retraining times compared to DL models when $\mA$ is changed, while matching their accuracy; although LSH schemes tended to have higher inference latencies, it was shown that the latency-gap could be bridged by building smaller databases with dimensionality-reduced features. Furthermore, we showed how the number of LSH databases and the hash size interplay to trade-off precision, recall and latency.

\begin{figure}[h]
    \centering
    \vspace{-0.5mm}
    \includegraphics[width=1.0\linewidth]{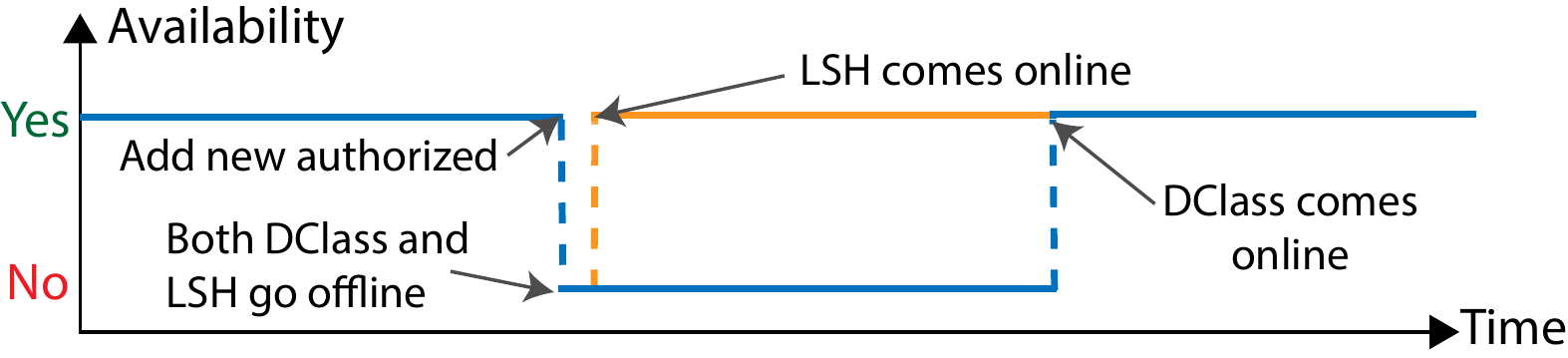}
    \caption{Using LSH-based authorization as backup for DClass}
    \label{fig:lsh_dclass_adapt_timeline}
\end{figure}

Even though we demonstrated many promising features of LSH-based authorization, these results are preliminary, and hence our message from this paper is not for them to replace DL models as the state-of-the-art. Since the LSH scheme we evaluated relied on a DL-based authenticator as its feature-extractor by design, our proposition is that they be used as a quick-adapting backup to DL models in the face of sudden changes in the authorized set: this is depicted in Fig. \ref{fig:lsh_dclass_adapt_timeline}. As the LSH scheme could be adapted quickly, we can use it as a backup authenticator while the DL model is down, and retrain the DL model in the background. This ensures that the authorization system experiences minimal downtime while not compromising much in terms of accuracy or latency.

\bibliographystyle{ieeetr}

\end{document}